\documentclass[10pt,twocolumn,letterpaper]{article}

\usepackage{iccv}
\usepackage{times}
\usepackage{epsfig}
\usepackage{graphicx}
\usepackage{amsmath}
\usepackage{amssymb}

\usepackage{multirow}
\usepackage{enumitem}
\usepackage{siunitx}
\usepackage{booktabs}
\usepackage[accsupp]{axessibility}
\usepackage{pdfpages}

\usepackage[breaklinks=true,bookmarks=false]{hyperref}

\iccvfinalcopy 

\def\iccvPaperID{****} 

\ificcvfinal\fi

\begin{document}

\title{Tracing the Influence of Predecessors on Trajectory Prediction}

\author{Mengmeng Liu, Hao Cheng, Michael Ying Yang\\
University of Twente, The Netherlands\\
{\tt\small  mengmeng.liu1998@gmail.com; \{h.cheng-2, michael.yang\}@utwente.nl}
}

\maketitle
\ificcvfinal\thispagestyle{empty}\fi

\begin{abstract}
   In real-world traffic scenarios, agents such as pedestrians and car drivers often observe neighboring agents who exhibit similar behavior as examples and then mimic their actions to some extent in their own behavior.
    This information can serve as prior knowledge for trajectory prediction, which is unfortunately largely overlooked in current trajectory prediction models.
    This paper introduces a novel Predecessor-and-Successor (PnS) method that incorporates a predecessor tracing module to model the influence of predecessors (identified from concurrent neighboring agents) on the successor (target agent) within the same scene. The method utilizes the moving patterns of these predecessors to guide the predictor in trajectory prediction.
    PnS effectively aligns the motion encodings of the successor with multiple potential predecessors in a probabilistic manner, facilitating the decoding process.
    We demonstrate the effectiveness of PnS by integrating it into a graph-based predictor for pedestrian trajectory prediction on the ETH/UCY datasets, resulting in a new state-of-the-art performance.
    Furthermore, we replace the HD map-based scene-context module with our PnS method in a transformer-based predictor for vehicle trajectory prediction on the nuScenes dataset, showing that the predictor maintains good prediction performance even without relying on any map information.
\end{abstract}

\section{Introduction}
\label{sec:intro}
\begin{figure}[hbpt!]
\begin{center}
 \includegraphics[clip=true, trim=0in 2.4in 3.2in 0in, width=\linewidth]{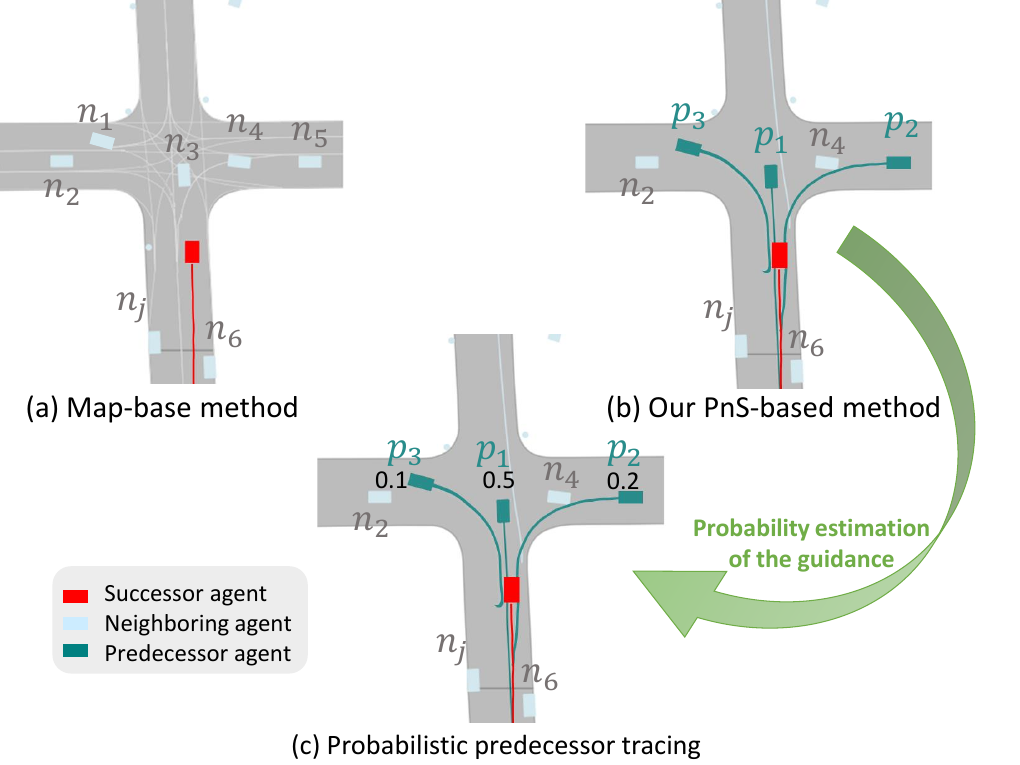}
\end{center}
  \caption{Compared to an HD-map based method (a) that relies on the lane segments for scene-aware trajectory prediction, our PnS-based method identifies predecessors, \ie $\{p_1, p_2, p_3\}$, from the neighboring agents, \ie $\{n_1, n_2, ..., n_j\}$, with similar moving patterns (b) and learns a probabilistic guidance of the predecessors influencing the successor agent (c). It should be noted that all the neighboring agents are still considered for interaction modeling in both methods.}
  \vspace{-5mm}
\label{fig:fig1}
\end{figure}
Trajectory prediction is a fundamental problem in various fields, such as transportation, robotics, and surveillance. 
Trajectory predictors, typically, rely on observed trajectories of the target and neighboring agents to model their motion dynamics and interactions~\cite{alahi2016social,lee2017desire,gupta2018social}.
In addition, state-of-the-art trajectory prediction methods, especially for autonomous driving, often make use of other prior information, such as semantic maps, \eg,~\cite{hong2019rules,salzmann2020trajectron++,phan2020covernet,yuan2021agentformer}, high-definition (HD) maps, \eg,~\cite{gao2020vectornet,khandelwal2020if,narayanan2021divide,zhou2022hivt,huang2022multi,wang2022ltp,zeng2021lanercnn,varadarajan2022multipath++,deo2022multimodal,park2023leveraging}, or goal information, \eg,~\cite{gilles2022gohome,gilles2021thomas,zhao2021tnt,gu2021densetnt,chiara2022goal}.

Besides that, in real-world traffic scenarios, we often observe that pedestrians and car drivers follow the trace of neighbors who have conducted a similar behavior in the same scene, assuming that all the agents behave rationally, \eg,~complaint to scene constraints and avoid collisions.
As shown in Fig.~\ref{fig:fig1}(b), given previously-captured frames in this toy scenario, the future trajectory of the target agent (red box) becomes more predictable if we can identify similar moving patterns in the same space that have demonstrated by other agents (dark green).
This implies that trajectories observed from other agents are not only useful for interaction modeling, but to some extent also can be treated as an informative prior to facilitate the prediction.

However, valuable dynamic information from the neighboring agents within the same scene to guide the target agent is often overlooked by mainstream trajectory prediction methods that heavily rely on static scene information, as shown in Fig.~\ref{fig:fig1}(a). 
Although existing memory-based search techniques like~\cite{marchetti2020multiple,xu2022remember,li2022graph} and clustering methods such as~\cite{sun2020recursive,xue2020poppl,meng2022forecasting} use historical trajectories to identify similar moving patterns, they require extra efforts to accumulate the historical trajectories before carrying out the searching step.
Also, there could be a huge time lag between the current target agent and the historical agents, which can seriously hinder the match of spatial and temporal relationships between the current and past traffic scenes which are associated with many changing factors.

Therefore, the research gap mentioned above motivates us to propose a novel Predecessor-and-Successor (PnS) method to explore the influence of neighboring agents (potential predecessors) on the guidance of the target agent (successor). 
Predecessors are defined as agents that have been concurrent in the same scene as the target agent and could potentially be part of its guide.
In other words, in addition to modeling the interactions between the target and neighboring agents for collision avoidance, we also explicitly model the potential guidance of the predecessors identified from these neighboring agents on the target agent.
As shown in Fig.~\ref{fig:fig1} (b) and (c), PnS employs an attention-based probabilistic approach to identify predecessors and learn their influence on the target agent. 
For example, neighboring agents $\{n_3, n_5, n_1\}$ are identified as potential predecessors $\{p_1, p_2, p_3\}$ indexed by the rank of probabilities based on the alignment between their spatial and temporal relationships.
Compared to the memory- and cluster-based methods, no extra efforts are needed to collect all the historical trajectories beforehand, and the time lag is much smaller from observing the predecessors to the successor following the predecessors' trace only after a few seconds. 

To demonstrate the effectiveness of the PnS method for both pedestrian and vehicle trajectory predictions, we integrate PnS into a state-of-the-art graph-based predictor GATraj \cite{cheng2022gatraj} for pedestrian trajectory prediction on the ETH/UCY~\cite{pellegrini2009you,lerner2007crowds} datasets, and we replace the HD map-based scene-context module with PnS in another transformer-based predictor LAformer \cite{liu2023laformer} that holds a high rank for vehicle trajectory prediction on the nuScenes~\cite{caesar2020nuscenes} benchmark. 
The \textbf{main contributions} of this work are summarized as follows:
    \begin{itemize}
    \item Our work proposes a novel Predecessor-and-Successor (PnS) method to learn the probabilistic influence from neighboring agents on the target agent. It effectively explores the predecessors motion as prior information to guide the prediction of the successor's trajectory.
    \item We demonstrate that leveraging predecessor information further pushes the state-of-the-art performance for the pedestrian trajectory prediction on the ETH/UCY datasets.
    In the mapless-based setting for the vehicle trajectory prediction on the nuScenes dataset, the PnS method largely mitigates the performance degradation when the HD-map information is removed.
\end{itemize}

\section{Related Work}
\label{sec:relatedwork}

\noindent
\textbf{Multimodal trajectory prediction.}
Due to the stochastic nature of road agents' future trajectories, arising from \eg, their randomness, subjectivity of intent, mutual influence, and scene constraints, predicting multimodal trajectories has become a prevalent approach in trajectory prediction. 
Here, multimodal prediction refers to generate multiple plausible trajectories of the target agent.
This trend is further motivated by large-scale benchmarks featuring real-world traffic scenarios~\cite{pellegrini2009you,lerner2007crowds,caesar2020nuscenes,chang2019argoverse,zhan2019interaction, Argoverse2,TrustButVerify}. 
Various methods address behavior uncertainty, including using Gaussian or Laplacian mixture models trained with Mixture Density Networks (MDNs) to estimate the likelihood of each mode~\cite{zhou2022hivt, salzmann2020trajectron++, gao2020vectornet, deo2022multimodal}. 
Another approach involves modeling multimodality implicitly via latent variables sampled from a prior distribution to generate diverse futures, including Variational Auto-Encoders (VAEs) \cite{salzmann2020trajectron++, yuan2021agentformer, chen2022scept}, Generative Adversarial Networks (GANs) \cite{gupta2018social, lee2017desire}, and Diffusion models \cite{gu2022stochastic,mao2023leapfrog}.
In this paper, we follow the mainstream approaches to demonstrate our PnS by integrating it into two representative multimodal trajectory predictors \cite{cheng2022gatraj,liu2023laformer}.
To account for the multimodality of predictions, PnS employs a predecessor tracing module to identify multiple potential predecessors and learn a probabilistic guidance from each of the predecessors on the target agent to generate multiple predictions.

\begin{figure*}[hbpt!]
\begin{center}
  \includegraphics[clip=true, trim=0in 3.1in 1.7in 0in, width=\linewidth]{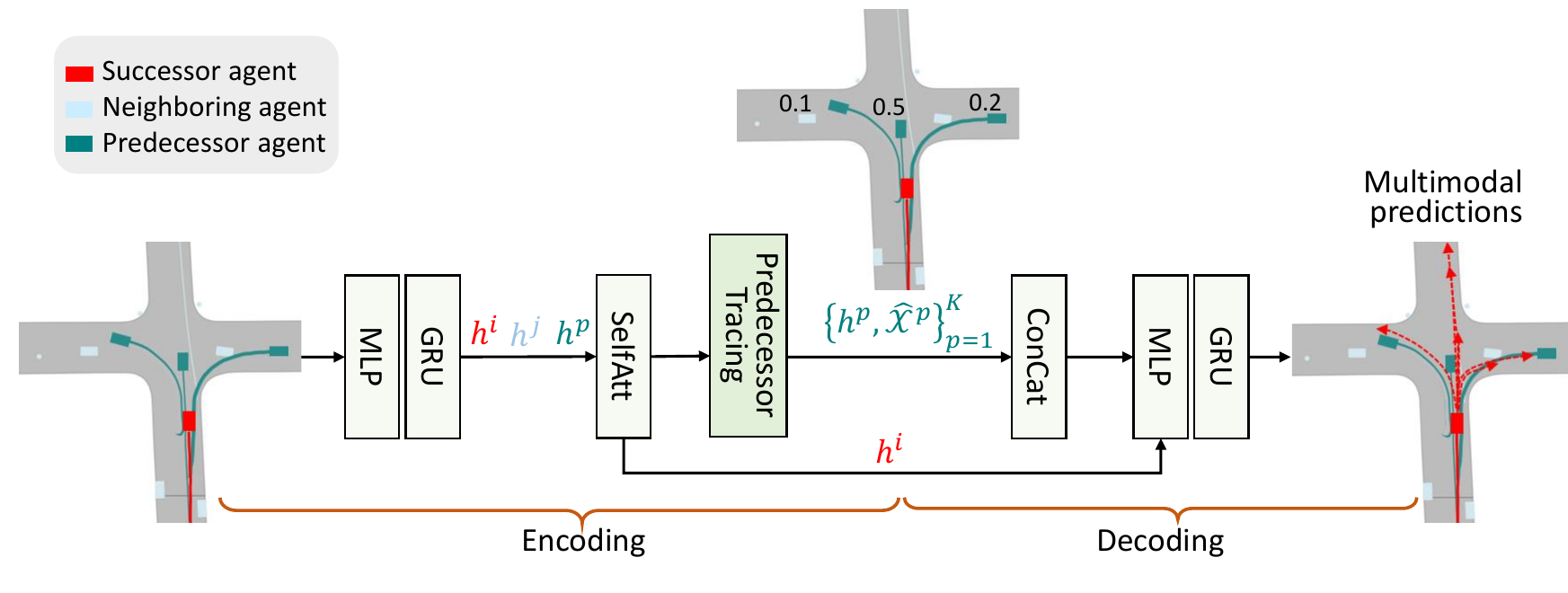}
\end{center}
\vspace{-4mm}
  \caption{The framework of the Predecessor-and-Successor (PnS)-based trajectory prediction model consists of several components. PnS utilizes a stack of Multi-Layer Perceptron (MLP), Gated Recurrent Unit (GRU), and self-attention layers to extract motion encodings $h^i$, $h^j$, $h^p$ of the successor, neighboring, and predecessor agents, respectively. The predecessor tracing module employs across-attention mechanisms to determine the probability of a predecessor, denoted as $p$, influencing the successor, denoted as $i$, among all the potential predecessors. This module learns the probabilities of predecessors' influence on the successor. To accommodate the stochastic behavior of the successor, the encoding of the top $K$ predecessors, based on the ranking of their probabilities, is aggregated with the encoding of the successor. This aggregation facilitates the decoding of multiple future trajectories for the successor.}
  \vspace{-2mm}
\label{fig:framework}
\end{figure*}

\vspace{1mm}
\noindent
\textbf{Interaction modeling.}
Effectively modeling the interactions among agents is crucial to account for their mutual influences.
Many of the approaches only focus on the interactions between the target and neighboring agents in the observation time for interaction modeling~\cite{sun2022m2i}.
Their motion encodings are aggregated by \eg,~pooling~\cite{alahi2016social,gupta2018social,xu2018encoding}, message passing~\cite{zhang2019sr,casas2020spagnn,li2022graph} via a multi-scale graph using Graph Convolutional Networks (GCNs)~\cite{welling2017semi}, and attention mechanisms~\cite{vaswani2017attention} that focus on the salient spatial and temporal features of the encodings~\cite{yuan2021agentformer,liu2021multimodal,chen2022scept,liu2023laformer,huang2022multi}.
Our work goes beyond by modeling the interactions between the target and neighboring agents in the observation time using the attention mechanisms.
We also condition the target agent's future trajectory on the historical trajectories by learning a probabilistic interactive influence of the predecessor agents. 
As we assume that these predecessors conducted their behavior rationally following the speed profiles complaint to scene constraints; The moving patterns derived from the predecessors in the same scene as the target agent is treated as a prior to better guide the prediction.  

\vspace{1mm}
\noindent
\textbf{Prior information from historical trajectories.}
Memory-based methods \cite{marchetti2020multiple,xu2022remember,li2022graph} utilize historical trajectories to improve prediction performance. 
For example, ~\cite{marchetti2020multiple} reads trajectories that are most likely to occur in future from stored observations in memory and uses them to augment the encodings of the target agent.
Similarly, \cite{xu2022remember} employs a pair of memory banks to store representative instances from the training set, acting as the prefrontal cortex in the neural system. 
It also employs a trainable memory addresser to adaptively search for situations similar to current ones in the memory bank, acting as the ``basal ganglia". 
\cite{li2022graph} forecasts multiple paths based on historical trajectories by modeling multi-scale graph-based spatial transformers combined with a trajectory smoothing algorithm named ``Memory Replay" using a memory graph. 
Moreover, cluster-based methods aggregate moving patterns, such as grouping target agent-neighbors~\cite{sun2020recursive, xue2020poppl} and focusing on historical group trajectories~\cite{meng2022forecasting}, for trajectory prediction. 
Also, \cite{sun2021three} clusters the trajectories into a fixed number of categories for trajectory classification.
However, there could be a huge time lag between the current and past traffic scenes that are associated with many changing factors, like traffic density and scene contexts.
This makes it difficult for the existing memory-based search techniques and clustering methods to distinguish the impact of historical trajectories on the target agent, taking into account its current traffic situation and motion dynamics. 
Moreover, these methods require to accumulate historical trajectories to build up the memory base beforehand, which causes extra burden and may limit their application in new scenes.
In contrast, we propose a probabilistic Predecessor-and-Successor method that leverages trajectories of the neighboring agents to identify representative predecessors as reference trajectories for the target agent.
Because the target and neighboring agents appear at the same time and share the same scene, their spatial and temporal relationships can be better modeled.  

\section{Method}
\label{sec:methodology}
\subsection{Problem formulation}
\label{subsec:problemformulation}
Following mainstream methods~\cite{gao2020vectornet,gu2021densetnt,zhou2022hivt,varadarajan2022multipath++}, we define the motion forecasting problem as predicting the subsequent trajectory $Y_{1:t_f}^i$ of a target agent $i$ given the set of observed trajectories $\mathbf{X}$ of a total of $N$ agents in the same scenario, including both the target and neighboring agents. 
To differentiate the target and neighboring agents, we use $X_{t_h-1:0}^i$ to denote the observed trajectory of target agent $i$ and $\hat{\mathbf{X}} = \mathbf{X} \setminus X^i$ the remaining set of the observed trajectories of neighboring agents.
The observation time horizon is $\{t_h-1,\dots, 0\}$ and the future time horizon is $\{1,\dots, t_f\}$.
To simplify the notation, time steps are omitted if they are not otherwise noted, and we call the target agent \textit{successor} in the remaining of this paper.

To fully exploit the observed trajectories of the neighboring agents $\hat{\mathbf{X}}$, a predecessor tracing function $g(\hat{\mathbf{X}}, X^i)$ learns a probabilistic guidance from the predecessors on the successor $i$.
Consequently, the prediction $\hat{Y}^i = f (\mathbf{X}, g(,))$ conditions $\hat{Y}$ on  $\mathbf{X}$ and $g(,)$.
Compared to many previous trajectory predictors formulated as $\hat{Y}^i = f (\mathbf{X})$, the only added component is $g(,)$, which does not require any extra input data.
Hence, this predecessor tracing can be easily incorporated into these predictors.

\subsection{Predecessor tracing}
\label{subsec:relation_predictor}
To effectively trace the impact of predecessors on the successor, we propose a time step-wise relationship predictor to identify their spatial relationship and use attention mechanisms to learn the impact between them. 

In order to model the influence from predecessors, we propose to use a cross-attention mechanism to design the interaction mapping function $\pi (\,,\,)$. 
Specifically, we use linear projections to transform predecessor $p$'s motion encoding $h^p$ into a query vector $Q$, and the successor's motion encoding $h^i$ into key ($K$) and value ($V$) vectors. 
These vectors are then used as inputs to a scaled dot-product attention block to calculate the attention score $s^{i,p}$.
\begin{equation}
    \label{eq:att}
    s^{i,p} = \operatorname{Softmax}(\frac{QK^{T}}{\sqrt{d_{k}}})V,
\end{equation}
where $d_k$ is the dimensionality of the key vectors. For more details on the encodings, refer to Sec.~\ref{subsec:prediction}.

The influence from the predecessor $p$ to successor $i$ is then mapped by taking as inputs the concatenation of $(h^i, h^p, s^{i,p})$.
Here, we implement a Multi-Layer Perceptron (MLP) for the mapping. 
The output of the MLP is denoted by ${\pi}(h^i, h^p)$.
\begin{equation}
    \label{eq:pi}
    {\pi}(h^i, h^p) = \operatorname{MLP}(h^i, h^p, s^{i,p}).
\end{equation}

Rather than using a deterministic approach to model the influence of predecessors, we propose a probabilistic approach that better considers the stochasticity of the successor's behavior. Specifically, we use a Softmax function to calculate the probability of predecessor $p$ influencing the successor $i$ at time step $t$ among all the potential $N \setminus i$ predecessors.
\begin{equation}
\label{eq:softmax_dense}
\hat{\mathcal{X}}_{t}^p = g(X^p, X^i) = \frac{\exp({\pi}(h_t^i, h_t^p))}{\sum^{N \setminus i}_{n=1} \exp({\pi}(h_t^i, h_t^n))}.
\end{equation}

\subsection{Trajectory encoding and decoding}
\label{subsec:prediction}
In this section, we describe the trajectory encoding and decoding process of our proposed model. 
Similar to previous attention-based models~\cite{yuan2021agentformer, liu2021multimodal, chen2022scept, liu2023laformer, huang2022multi}, we utilize the self-attention mechanism to encode the motion dynamics of the successor, neighboring, and predecessor agents, and decode the future trajectories of the successor, as shown in Fig.~\ref{fig:framework}.

\vspace{6pt}
\noindent
\textbf{Encoding.} 
We start by extracting the motion encodings of the successor, neighboring, and predecessor agents using a MLP followed by a Gated Recurrent Unit (GRU).
\begin{equation}
\label{eq:encoding}
    h_{t}^{o} = \operatorname{GRU}( \operatorname{MLP}(X_{t}^{o})),
\end{equation}
where $o = \{i,\,j,\,p\}$ denotes the index of the successor, neighboring, or predecessor agent, and $j, p \in N \setminus i$.

Next, we aggregate the interaction information among these three types of agents using the self-attention mechanism, as shown in Eq.~\eqref{eq:interaction}:
\begin{equation}
\label{eq:interaction}
    h_{t}^{o} = h_{t}^{o} + \operatorname{SelfAtt}( h_{t}^{o}),
\end{equation}
where $\operatorname{SelfAtt}$ represents the self-attention function.
In this step, the pair-wise interaction information among them is learned by the attention and added to their original encodings via the skip connection.
In this way, their interconnections are aggregated.

\vspace{6pt}
\noindent
\textbf{Decoding.}
To generate multimodal predictions for the successor, we implement a Laplacian Mixture Density Network (MDN) decoder following \cite{zhou2022hivt,liu2023laformer}.

First, instead of focusing on the impact of a single predecessor, we include multiple potential predecessors to enable the decoder to mimic the successor's stochastic behavior. 
For example, the predecessors may turn into different directions at an intersection or drive at different speeds.
To achieve this goal, we select $K$ predecessors based on their probability score $\hat{\mathcal{X}}_{t}^p$ in a descending order and aggregate these predecessors' motion encodings, as shown in Eq.~\eqref{eq:ancestoragg}:
\begin{equation}
    \label{eq:ancestoragg}
    \hat{\mathcal{X}}_{t}^K = \operatorname{ConCat}[\{(h_t^p, \hat{\mathcal{X}}_{t}^p)\}_{p=1}^K].    
\end{equation}
After that, the decoder takes as input the motion encoding $h_{t}^{i}$ of the successor and $\hat{\mathcal{X}}_{t}^K$ of the $K$ most influencing predecessors and outputs the distributions of the successor's future positions.
More specifically, The outputs are parameterized by the location $\mu^{m}$ and scale ${b}^{m}$ parameters of a total of $M$ components, where $m \in M$ and $M$ corresponds to the different modalities of the predictions.

Similar to the encoding process, we use a MLP layer followed by a GRU layer to implement the decoder. It outputs the location and scale parameters of each component of the MDN, as well as the associated likelihood $\pi_{m}$. 

\subsection{Incorporating with the existing models}
To analyze the effectiveness of the proposed PnS method, we incorporated it into two trajectory predictors -- GATraj \cite{cheng2022gatraj} and LAformer \cite{liu2023laformer}.
These two models are chosen based on the following reasons: 
First, both GATraj and LAformer are one of the latest models and have shown excellent performances on the ETH/UCY datasets for pedestrian trajectory prediction and nuScenes dataset for vehicle trajectory prediction, respectively; 
Second, they apply different mainstream frameworks for trajectory prediction. Namely, GATraj exploits a graph-based framework with massage passing, while LAformer employs the transformer framework with attention mechanisms, to learn spatial and temporal information for trajectory prediction. 
These two different models are representative examples to demonstrate the compatibility of the PnS method.
Third, both models use a MDN decoder, the PnS method can be smoothly incorporated into the backbones and jointly trained without drastic modification. 
It should be noted that because the original LAformer heavily relies on lane segments from an HD map as a strong prior to guide the prediction. 
Hence, we substitute the lane alignment module in LAformer with our PnS method to guide the prediction.  

\subsection{Training}
\label{sub:loss}
The predecessor tracing module is optimized using the binary cross-entropy loss $\mathcal{L}_\text{PnS}$ to improve the probability estimation. The loss function is defined as follows:
\begin{equation}
\label{eq:loss_lane}
\mathcal{L}_\text{PnS} = \sum^{t_\text{f}}_{t=1}\mathcal{L}_\text{CE}(\mathcal{X}_t^p ,\hat{\mathcal{X}}_t^p ),
\end{equation}
where $\mathcal{X}_t^p$ and $\hat{\mathcal{X}}_t^p$ denote the ground truth and predicted probability of predecessor $p$ influencing the successor $i$ at time step $t$, respectively.

In the training phase, we identify the predecessor that has the closest spatial relationship and moving patterns with the successor at each time step. 
This is achieved by identifying predecessor $p$ from $\mathbf{\hat{X}}$ using a distance metric $\phi$.
\label{eq:ancestor}
\[ \mathcal{X}_t^p =
  \begin{cases}
    1       & \quad \text{if } p = \underset{{p\in N \setminus i}}{\operatorname{arg\,min}}~\phi (Y_t^i, \hat{X}),\\
    0  & \quad \text{otherwise}.
  \end{cases}
\]
Specifically, we use the $L_2$ distance metric to find the predecessor that is closest to the successor at time step $t$. 
The identified predecessor is labeled as the true predecessor, while all other agents from $\mathbf{\hat{X}}$ are labeled as false predecessors.

Following previous works~\cite{makansi2019overcoming,zhou2022hivt,deo2022multimodal}, we optimize the decoder in both GATraj and LAformer using the negative log-likelihood (NLL) of the best predicted mode $m^*$ of the Laplacian MDN. 
\begin{equation}
 \mathcal{L}_{\text{NLL}}=\frac{1}{t_{f}}\sum^{t_{f}}_{t=1}-\log P({Y_{t}}|\mu^{m^*}_{t}, \mathbf{b}^{m^*}_{t}),
\end{equation}
where $\mu^{m^*}$ and ${b}^{m^*}$ are the location and scale parameters of the component, respectively, and $m^*$ represents the mode with the minimum $L_2$ error of the predicted and ground truth trajectories among the total $M$ components.
We utilize the cross-entropy to optimize the mode classification.
\begin{align}
\mathcal{L}_{\text{cls}}=\sum^{M}_{m=1}-\pi_{m}\log(\hat{\pi}_{m}),
\end{align}
where $\pi_{m}$ denotes the target probability of the mode.
This target probability is defined by a soft displacement function, using the same method proposed in~\cite{zhou2022hivt}.

The overall objective function is formulated as:
\begin{equation}
\label{eq:loss}
\mathcal{L} =  \lambda \mathcal{L}_{\text{PnS}} + \mathcal{L}_{\text{cls}} + \mathcal{L}_{\text{NLL}} ,
\end{equation}
where $\lambda $ is a hyperparameter that controls the weights of loss terms in the objective function, allowing us to balance their respective contributions.

\section{Experiments}
\label{sec:experiments}
\subsection{Experimental setup}
\label{subsec:setup}
\noindent
\textbf{Dataset.} 
We utilize the ETH/UCY datasets~\cite{pellegrini2009you,lerner2007crowds} to train and test our PnS method using GATraj \cite{cheng2022gatraj} for pedestrian trajectory prediction. 
These datasets consist of multiple subsets captured at different locations, each with varying pedestrian densities, including \textit{Eth}, \textit{Hotel}, \textit{Uni}, \textit{Zara1}, and \textit{Zara2}. 
Following the most common setting \cite{alahi2016social,gupta2018social,salzmann2020trajectron++,yuan2021agentformer,meng2022forecasting}, the trajectories are down-sampled to a frequency of \SI{2.5}{Hz}, with an observation time horizon of \SI{3.2}{s} and a prediction time horizon of \SI{4.8}{s}.
For training and testing, we follow the standard leave-one-out data partitioning to train the models on four out of these five subsets and test them on the remaining one. 
This procedure is repeated for each subset.
Moreover, we employ the nuScenes dataset \cite{caesar2020nuscenes} for training and testing our PnS method using LAformer \cite{liu2023laformer} for vehicle trajectory prediction. This dataset encompasses various driving scenarios involving complex intersections and interactions with pedestrians, cyclists, and other elements. It comprises a total of 245,414 trajectory instances across 1,000 driving scenes, each lasting 20 seconds and sampled at 2Hz. The observation time horizon is set to two seconds, while the prediction time horizon is set to six seconds.
To facilitate offline training and validation, 850 scenes are provided with ground truth information, while the remaining 150 scenes are reserved for online testing.

\vspace{1mm}
\noindent
\textbf{Evaluation metrics.} 
We adhere to standard evaluation metrics to assess the prediction performance \cite{alahi2016social,caesar2020nuscenes,zhou2022hivt}. 
Specifically, we employ displacement errors (DE) to evaluate the performance on the ETH/UCY and nuScenes datasets.
In particular, we utilize two metrics: $\text{mFDE}\mathsf{K}$ and $\text{mADE}_\mathsf{K}$, which measure the minimum $L_2$ errors in meters at the Final step and the Average of each step, respectively, for predicting $\mathsf{K}$ modes. 
The letter ``m" signifies the minimum error among the $\mathsf{K}$ modes. 
In the case of nuScenes, we set $\mathsf{K}$ to five or ten, while for ETH/UCY, it is set to 20.
It is important to note that for all evaluation metrics, a lower value indicates better performance.

\vspace{1mm}
\noindent
\textbf{Implementation details.} 
To ensure compatibility between PnS and GATraj \cite{cheng2022gatraj} and LAformer \cite{liu2021multimodal}, we set the hidden dimensions of all feature vectors to match the original configuration.
We set the predecessor number $K$ to 2, aiming to strike a balance between the number of predecessors and the diversity of predictions.
The value of $\lambda$ is set to 0.5, which helps maintain a balance among the loss terms in the objective function indicated by Eq.~\eqref{eq:loss}.
For the activation function of the intermediate layers, we use the Rectified Linear Unit (ReLU).
In the decoder, we employ the Exponential Linear Unit (ELU), specifically $\text{ELU}(.) + 1 + \epsilon$, as the activation function to generate positive probability estimations. Here, $\epsilon$ is set to $1e^{-3}$.
All the models were trained on eight RTX 3090 GPUs using the Adam optimizer \cite{kingma2015adam}.

\subsection{Quantitative results and comparison}
\label{subsec:results}
To effectively demonstrate the performance enhancement achieved by incorporating the PnS method into GATraj and LAformer, we present a comparison of their performance with and without PnS in Tables \ref{tab:ethucypns} and \ref{tab:nuscenepns}.

Table \ref{tab:ethucypns} reveals that GATraj+PnS exhibits improved performance on Eth, Uni, and Zara2, as measured by $\text{FDE}_\text{20}$. Despite GATraj already yielding minimal errors in each subset, the addition of our PnS method further reduces prediction errors, particularly in datasets with high pedestrian density such as Uni by 2.6\% in $\text{mFDE}_\text{20}$ and Zara2 by 8.3\% and 4.8\% in $\text{mADE}_\text{20}$ and $\text{mFDE}_\text{20}$, respectively. 
This is because the predecessor tracing module can derive information cues from more neighboring agents.
\begin{table}[hbpt!]
\centering
\footnotesize
\setlength{\tabcolsep}{2.5pt}
\begin{tabular}{l|c|c|c|c|c}
\toprule
Models           & Eth       & Hotel     & Uni       & Zara1     & Zara2      \\ \midrule 
GATraj\,\cite{cheng2022gatraj} & 0.26/0.42 & {0.10}/{0.15} & 0.21/0.38 & 0.16/{0.28} & 0.12/0.21   \\
GATraj\,\cite{cheng2022gatraj}+PnS & 0.26/{0.40} & {0.10}/{0.15} & 0.21/{0.37} & 0.16/{0.28} & {0.11}/{0.20} \\
Improvement & -/{\color{blue}4.7\%} & -/- & -/{\color{blue}2.6\%} & -/-& {\color{blue}8.3\%}/{\color{blue}4.8\%} \\ \bottomrule
\end{tabular}
\caption{Quantitative results on ETH/UCY \cite{pellegrini2009you,lerner2007crowds} measured by $\text{mADE}_\text{20}$/$\text{mFDE}_\text{20}$.}
\label{tab:ethucypns}
\end{table}

Table \ref{tab:nuscenepns} showcases the performance of PnS integrated into LAformer on the nuScenes dataset.
The comparison between LAformer and LAformer+PnS reveals that LAformer's performance is significantly compromised when it relies solely on observed trajectories without any map information. 
However, the inclusion of the PnS method greatly mitigates this issue, resulting in a reduction of approximately 5.1\% in mADE$_{5}$ and 9.8\% in mADE$_{10}$. 
This advantage indicates that the PnS method offers a practical alternative for vehicle trajectory prediction in scenarios where map information is unavailable, such as when a vehicle enters a new location without access to HD map data.

Interestingly, we also observe that providing both HD map and predecessor information to LAformer does not yield a combined improvement. 
Our conjecture is that the HD map already contains rich contextual details, including lane segments, road geometry, and traffic rules. 
Consequently, when we simply concatenate these two types of information, the PnS method may not exert a strong influence on the successors in this specific setup.
\begin{table}[ht!]
\centering
\footnotesize
\begin{tabular}{l|cc}
\toprule
Model  & mADE$_{5}$ &  mADE$_{10}$  \\ \midrule
LAformer\,\cite{liu2021multimodal} & 1.57 & 1.32  \\
LAformer\,\cite{liu2021multimodal}+PnS & {1.49}& {1.19}\\ 
Improvement & {\color{blue}5.1\%} & {\color{blue}9.8\%} \\ \midrule
LAformer\,\cite{liu2021multimodal}+HD & {1.19} & {0.93} \\ 
LAformer\,\cite{liu2021multimodal}+HD+PnS & 1.20 & {0.93} \\ 
Improvement & {\color{green}-0.8\%} & - \\ 
\bottomrule
\end{tabular}
\caption{The results on the \textit{nuScenes} \cite{caesar2020nuscenes} test set.}
\label{tab:nuscenepns}
\end{table}

Furthermore, we conduct a benchmark comparison of GATraj+PnS and LAformer+PnS with current models in Tables \ref{tab:resultsonethucy} and \ref{tab:resultsonnuscenes} for pedestrian and vehicle trajectory prediction, respectively.

In Table \ref{tab:resultsonethucy}, we compare GATraj+PnS with the most recent top models. Among the other models, particularly the Retrospective-Memory-based \textit{MemoNet} \cite{xu2022remember} and the scene history-based \textit{SHENet} \cite{meng2022forecasting} share similarities with the PnS concept to use existing trajectories as prior information. We also include the latest diffusion-based \textit{LED} \cite{mao2023leapfrog} that holds the top performance for this task. 
GATraj+PnS achieves the best performance across all subsets, except for Zara1 in terms of $\text{FDE}_\text{20}$ compared to \textit{LED}. 
This sets a new state-of-the-art performance on the ETH/UCY datasets for pedestrian trajectory prediction, even without the effort to accumulate historical trajectories for building up a memory base.
\begin{table}[hbpt!]
\vspace{-3mm}
\footnotesize
\centering
\setlength{\tabcolsep}{2.2pt}
\begin{tabular}{l|c|c|c|c|c}
\toprule
Models                       & Eth       & Hotel     & Uni       & Zara1     & Zara2   \\ \midrule 

Social-GAN\cite{gupta2018social}      & 0.81/1.52 & 0.72/1.61 & 0.60/1.26 & 0.34/0.69 & 0.42/0.84  \\
\footnotesize{Trajectron++}\cite{salzmann2020trajectron++} & 0.67/1.18 & 0.18/0.28 & 0.30/0.54 & 0.25/0.41 & 0.18/0.32\\
STAR\,\cite{yu2020spatio} & 0.36/0.65 & 0.17/0.36 & 0.31/0.62 & 0.26/0.55 & 0.22/0.46 \\
AgentFormer \cite{yuan2021agentformer} & 0.45/0.75 & 0.14/0.22 & 0.25/0.45 & 0.18/0.30 & 0.14/0.24 \\
MID\,\cite{gu2022stochastic}  & 0.39/0.66 & 0.13/0.22 & 0.22/0.45 & 0.17/0.30 & 0.13/0.27  \\
LB-EBM\,\cite{pang2021trajectory} & 0.30/0.52 & 0.13/0.20 & 0.27/0.52 & 0.20/0.37 & 0.15/0.29 \\
PCCSNet\,\cite{sun2021three} & 0.28/0.54 & 0.11/0.19 & 0.29/0.60 & 0.21/0.44 & 0.15/0.34 \\
GP-Graph\,\cite{bae2022learning} & 0.43/0.63 & 0.18/0.30 & 0.24/0.42 & 0.17/0.31 & 0.15/0.29 \\
MemoNet\,\cite{xu2022remember}      & 0.40/0.61 & 0.11/0.17 & 0.24/0.43 & 0.18/0.32 & 0.14/0.24  \\ 
SHENet\,\cite{meng2022forecasting}      & 0.41/0.61 & 0.13/0.20 & 0.25/0.43 & 0.21/0.32 & 0.15/0.26  \\ 
LED\,\cite{mao2023leapfrog}         & 0.39/0.58 & 0.11/0.17 & 0.26/0.43 & 0.18/\textbf{0.26} & 0.13/0.22  \\ 
GATraj\,\cite{cheng2022gatraj}+PnS        & \textbf{0.26}/\textbf{0.40} & \textbf{0.10}/\textbf{0.15} & \textbf{0.21}/\textbf{0.37} & \textbf{0.16}/0.28 & \textbf{0.11}/\textbf{0.20}  \\\bottomrule
\end{tabular}
\caption{The comparison on the ETH/UCY datasets measured by $\text{ADE}_\text{20}$/$\text{FDE}_\text{20}$. The best performance is in boldface.}
\vspace{-3mm}
\label{tab:resultsonethucy}
\end{table}

We compare LAformer with PnS to recent models on the nuScenes dataset. 
To ensure a fair comparison, we ensure that all models utilize the same input data.
First, we compare LAformer+PnS with AgentFormer \cite{yuan2021agentformer} and SG-Net \cite{wang2022stepwise}, as all these models solely rely on observed trajectories without incorporating map information.
Next, we compare PnS with LBA \cite{zhong2022aware}, which utilizes a Local Behavior-Aware module to leverage historical trajectories that have previously traversed the scene. 
This module bears resemblance to PnS in the way it explores existing trajectories to guide predictions.
Additionally, since the backbone models of LBA\footnote{We were unable to implement LBA with LAformer and directly compare it with LAformer+PnS+HD due to the broken link to the code of \cite{zhong2022aware}.}, namely P2T \cite{deo2020trajectory} and LaneGCN \cite{liang2020learning}, also incorporate HD map information, we compare them with LAformer+PnS+HD.
As demonstrated in Table \ref{tab:resultsonnuscenes}, LAformer+PnS exhibits significantly better performance compared to AgentFormer and SG-Net when the HD map is removed. 
In the map-based setting, LAformer+PnS+HD outperforms P2T+LBA+HD and achieves comparable performance to LaneGCN+LBA+HD.
It is worth mentioning that P2T+LBA+HD and LaneGCN+LBA+HD require access to historical trajectories, which necessitates pre-accumulation of these trajectories. 
In contrast, P2T+LBA+HD directly traces predecessors from concurrent neighboring agents within the same scene without extra input data in a more challenging yet practical setting.

\begin{table}[bpht!]
\centering
\footnotesize
\setlength{\tabcolsep}{1pt}
\begin{tabular}{l|ccc}
\toprule
Model & mADE$_5$ &  mADE$_{10}$ & mFDE$_1$ \\ \midrule
AgentFormer\,\cite{yuan2021agentformer} & 1.97  &  1.58 & - \\
SG-Net\,\cite{wang2022stepwise}& 1.86 &  1.40 & - \\
LAformer\,\cite{liu2021multimodal}+PnS & \textbf{1.49} & \textbf{1.19} & \textbf{8.08}\\ \midrule
P2T\,\cite{deo2020trajectory}+LBA\,\cite{zhong2022aware}+HD & - & 1.08 & 9.25 \\ 
LaneGCN \cite{liang2020learning}+LBA\,\cite{zhong2022aware}+HD & - & 0.95 & \textbf{6.78} \\
LAformer\,\cite{liu2021multimodal}+PnS+HD & \textbf{1.20} & \textbf{0.93} &  6.99 \\ \bottomrule
\end{tabular}
\caption{The prediction comparison on the nuScenes \cite{caesar2020nuscenes} test set. The best values are highlighted in boldface.}
\label{tab:resultsonnuscenes}
\end{table}

\subsection{Ablation study}
\label{subsec:ablation}
To analyze the effectiveness of the proposed components in the PnS method, we conduct an ablation study using the train-val split provided by the nuScenes dataset. 

Firstly, we ablate the predecessor tracing (PT) module to evaluate its efficacy in our proposed model, denoted as PnS-Baseline. 
In this setting, the prediction model can only condition future trajectories on the observed trajectories.
As shown in Table~\ref{tab:abanp}, removing the PT module results in a significant drop in performance, about 5\% in both $\text{mADE}_5$ and $\text{mFDE}_5$ for predicting five trajectories, and approximately 11\% in $\text{mADE}_{10}$ and 17\% in $\text{mFDE}_{10}$ for predicting ten trajectories.
\begin{table}[ht]
\centering
\footnotesize
\begin{tabular}{c|cccc}
\toprule
PT &  $\text{mADE}_5$  & $\text{mFDE}_5$&  $\text{mADE}_{10}$  & $\text{mFDE}_{10}$   \\ \midrule
$\surd$  & 1.49 &  3.10 & 1.19 &  2.24\\
 - & 1.57 &  3.28 & 1.32 &  2.63 \\ \bottomrule
\end{tabular}
\caption{Ablation study on PnS.} 
\vspace{-2mm}
\label{tab:abanp}
\end{table}

Secondly, we analyze the number of predecessors aggregated via Eq.~(\ref{eq:ancestoragg}) for the predecessor tracing module, as shown in Table~\ref{tab:topk}.
With a single predecessor, the model performs slightly worse than with two predecessors because the successor can be influenced by other predecessors as well. 
However, increasing the number of potential predecessors to three leads to a clear performance drop, as measured by $\text{mFDE}_{10}$. 
This is because the chances of including less influential or even irrelevant predecessors also increase.
\begin{table}[bpht!]
\centering
\footnotesize
        \begin{tabular}{l|cc}
        \toprule
        $k$ &  $\text{mADE}_{10}$  & $\text{mFDE}_{10}$   \\ \midrule
        1 &1.19 &2.27\\
        \underline{2}        & 1.19   & 2.24 \\ 
        3        & 1.21 &  2.32  \\ \bottomrule       
        \end{tabular}
\caption{The number of top predecessors.} 
\label{tab:topk}
\end{table}

Thirdly, we explore another distance metric in $\phi$ for predecessor identification.
Namely, we substitute $L_2$ by $L_1$ in $\phi$.  
As shown in Table~\ref{tab:phi}, $L_1$ does not work as well as $L_2$. 
It increases, \eg.~$\text{mFDE}_{10}$, by about 3\%. 
\begin{table}[bpht!]
\centering
\footnotesize
        \begin{tabular}{l|cc}
        \toprule
        Distance &  $\text{mADE}_{10}$  & $\text{mFDE}_{10}$   \\ \midrule
        $L_{1}$        & 1.24   & 2.43 \\ 
        \underline{$L_{2}$}        & 1.22 &  2.36  \\ \bottomrule       
        \end{tabular}
\caption{Distance metric for predecessor identification.} 
\label{tab:phi}
\vspace{-4mm}
\end{table}

Finally, we investigate the sensitivity of the weight parameter $\lambda$ in the loss function presented in Eq.~\eqref{eq:loss}. 
We vary the value of $\lambda$ while keeping other settings unchanged to analyze its effect on the prediction performance. 
We observe that decreasing or increasing this value to 0.1 or 1 leads to slightly worse performance in terms of $\text{mADE}_{10}$ and $\text{mFDE}_{10}$, as shown in Table \ref{tab:lambda}.
\begin{table}[bpht!]
\centering
\footnotesize
        \begin{tabular}{l|cc}
        \toprule
         $\lambda $ &  $\text{mADE}_{10}$  & $\text{mFDE}_{10}$   \\ \midrule
        0.1 & 1.25& 2.45\\
        \underline{0.5}        & 1.22   & 2.36  \\ 
        1        & 1.28 &  2.53   \\ \bottomrule       
        \end{tabular}
\caption{Sensitivity analysis of $\lambda $ in the objective function.} 
\label{tab:lambda}
\vspace{-4mm}
\end{table}

\begin{figure*}
    \centering
    \includegraphics[clip=true, trim=0pt 4.2cm 25.7cm 2.8cm, width=0.245\textwidth]{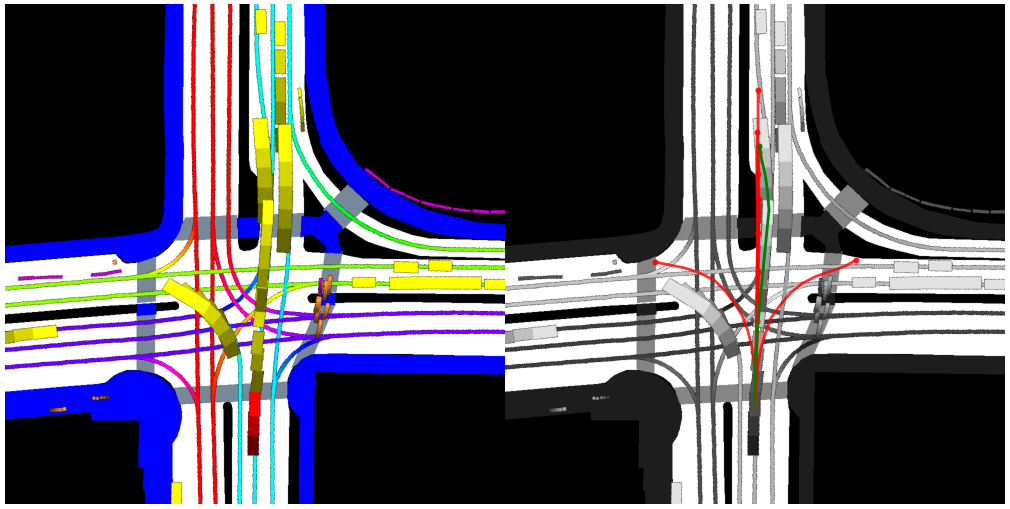}
    \includegraphics[clip=true, trim=25.7cm 4.2cm 0pt 2.8cm, width=0.245\textwidth]{Ancestor_fig/noanp_5.png}
    \includegraphics[clip=true, trim=25.7cm 4.2cm 0pt 2.8cm, width=0.245\textwidth]{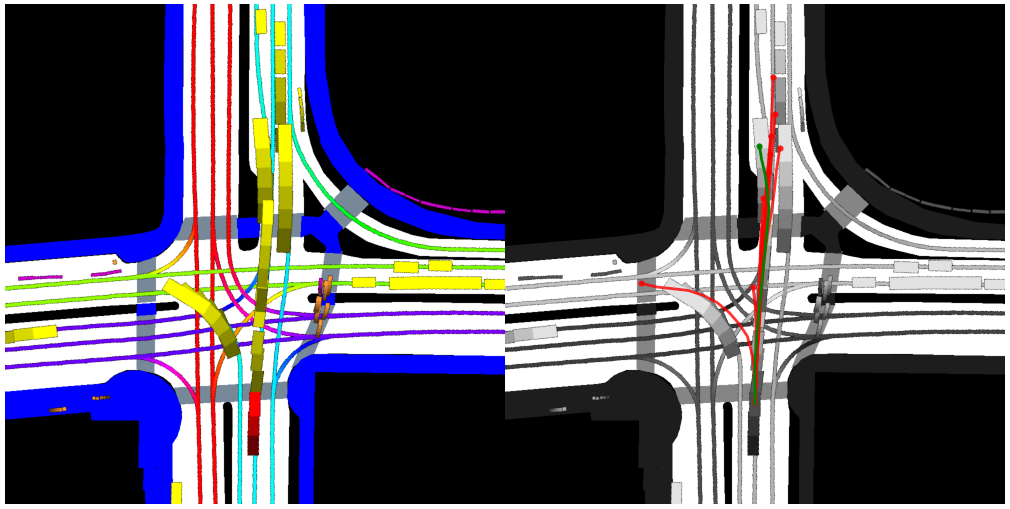} 
    \includegraphics[clip=true, trim=25.7cm 4.2cm 0pt 2.8cm, width=0.245\textwidth]{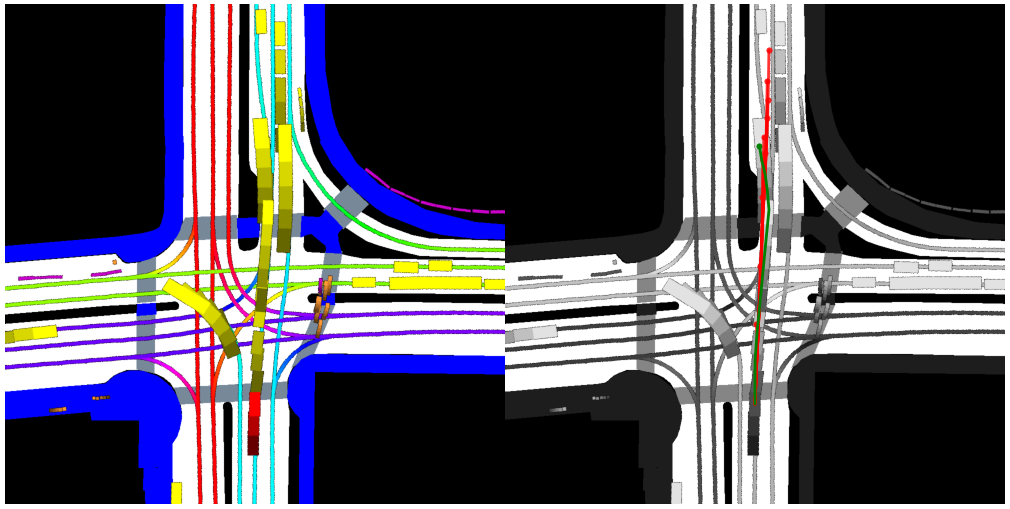}

    \includegraphics[clip=true, trim=0pt 3cm 25.7cm 5cm, width=0.245\textwidth]{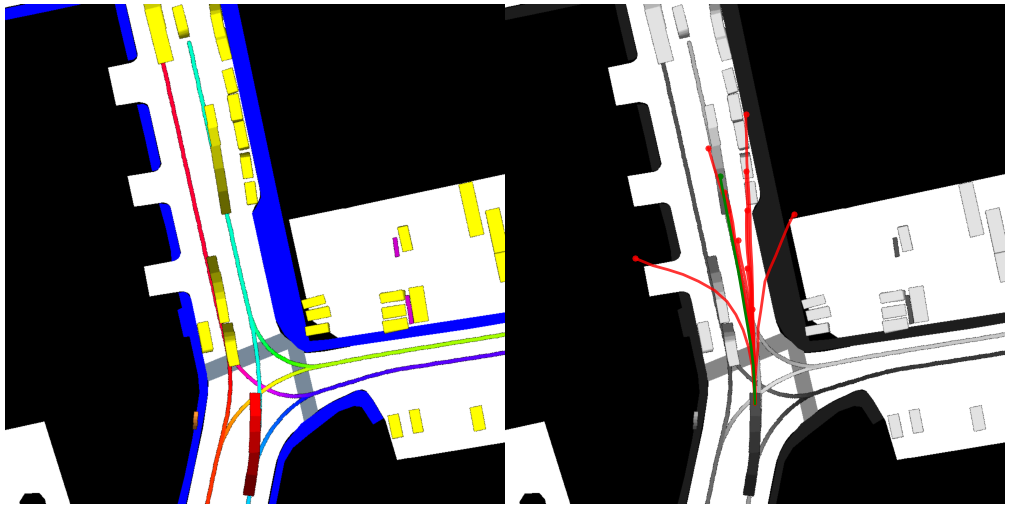}
    \includegraphics[clip=true, trim=25.7cm 3cm 0pt 5cm, width=0.245\textwidth]{Ancestor_fig/noanp_2.png}
    \includegraphics[clip=true, trim=25.7cm 3cm 0pt 5cm, width=0.245\textwidth]{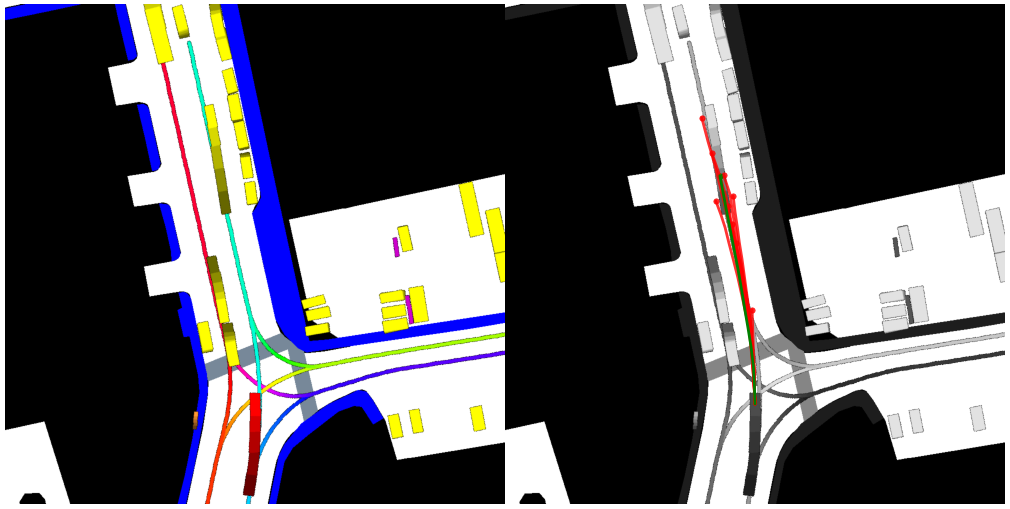} 
    \includegraphics[clip=true, trim=25.7cm 3cm 0pt 5cm, width=0.245\textwidth]{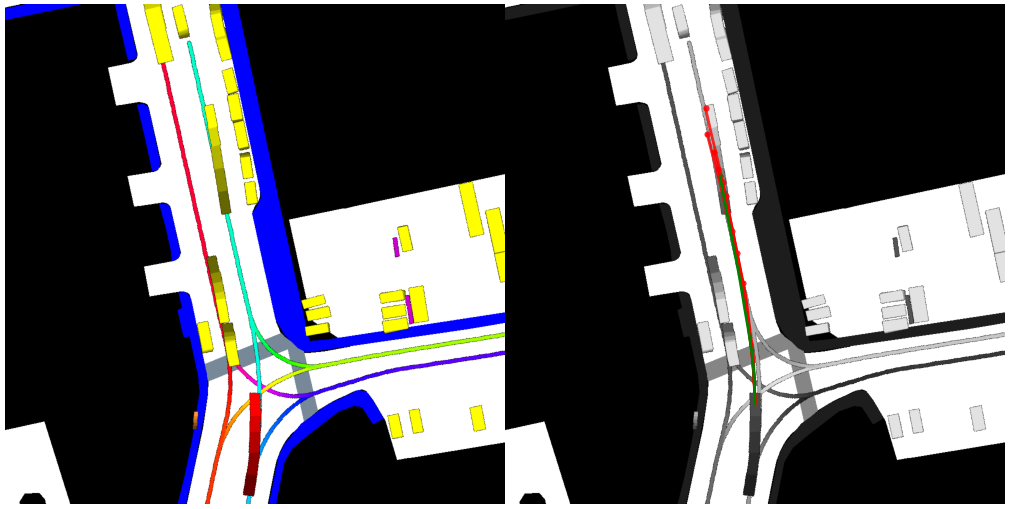}

    \includegraphics[clip=true, trim=0pt 4cm 25.7cm 5cm, width=0.245\textwidth]{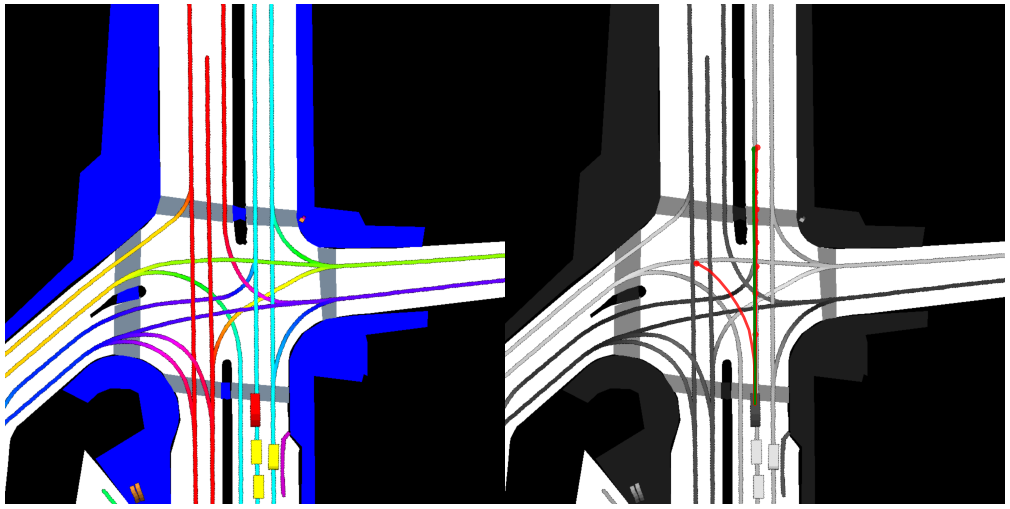}
    \includegraphics[clip=true, trim=25.7cm 4cm 0pt 5cm, width=0.245\textwidth]{Ancestor_fig/noanp_3.png}
    \includegraphics[clip=true, trim=25.7cm 4cm 0pt 5cm, width=0.245\textwidth]{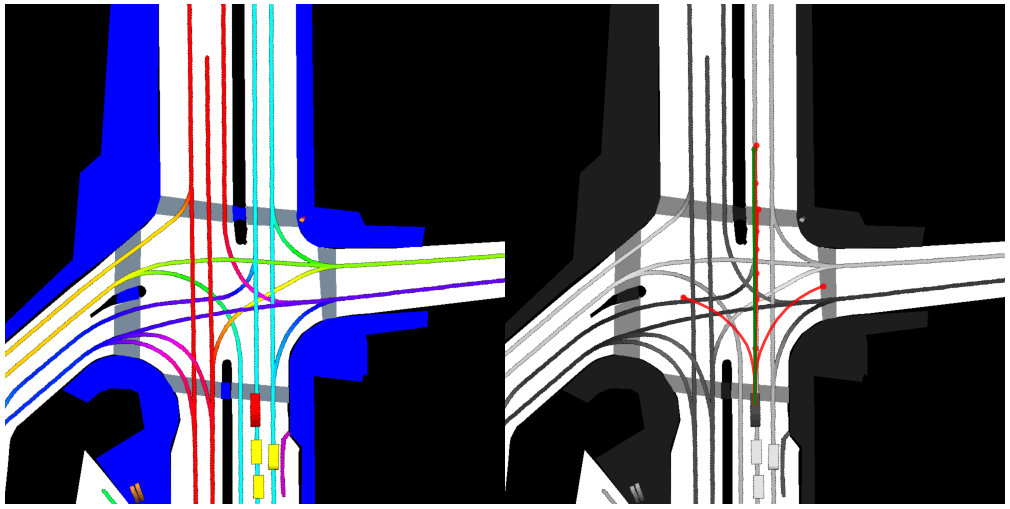} 
    \includegraphics[clip=true, trim=25.7cm 4cm 0pt 5cm, width=0.245\textwidth]{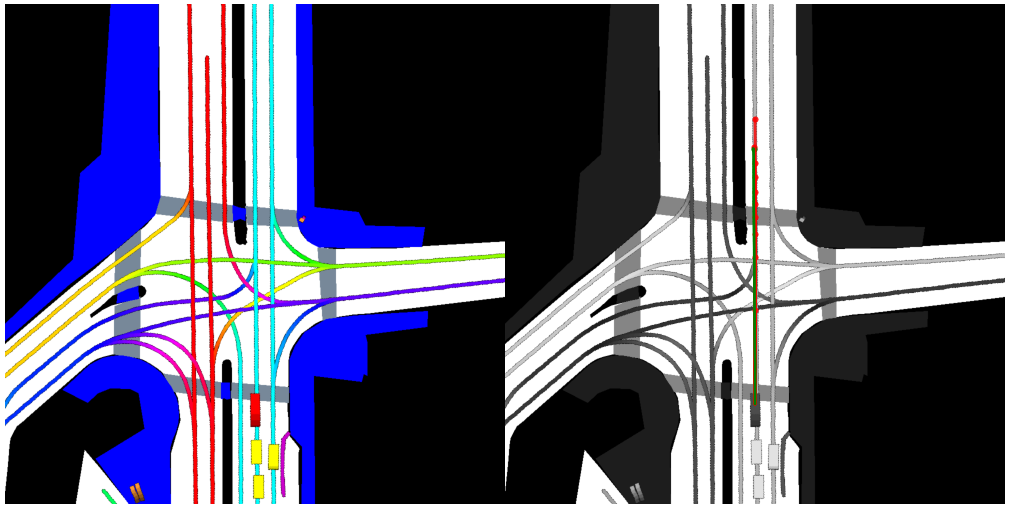}    

   \includegraphics[clip=true, trim=0pt 4cm 25.7cm 5cm, width=0.245\textwidth]{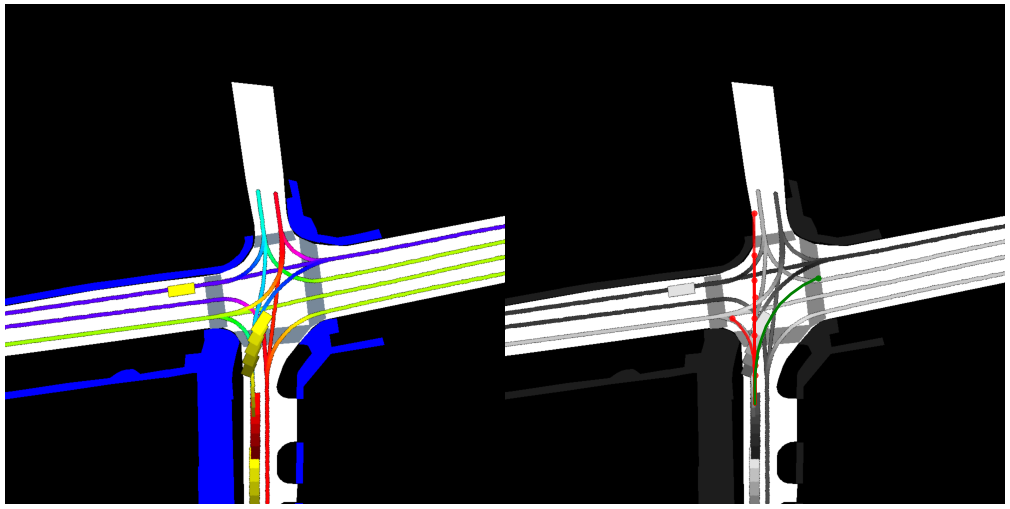}
    \includegraphics[clip=true, trim=25.7cm 4cm 0pt 5cm, width=0.245\textwidth]{Ancestor_fig/noanp_6.png}
    \includegraphics[clip=true, trim=25.7cm 4cm 0pt 5cm, width=0.245\textwidth]{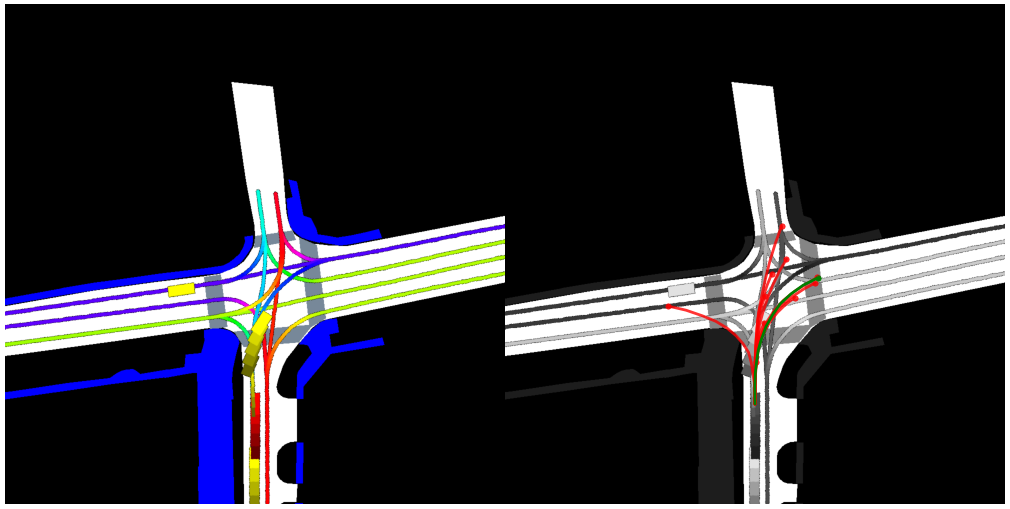} 
    \includegraphics[clip=true, trim=25.7cm 4cm 0pt 5cm, width=0.245\textwidth]{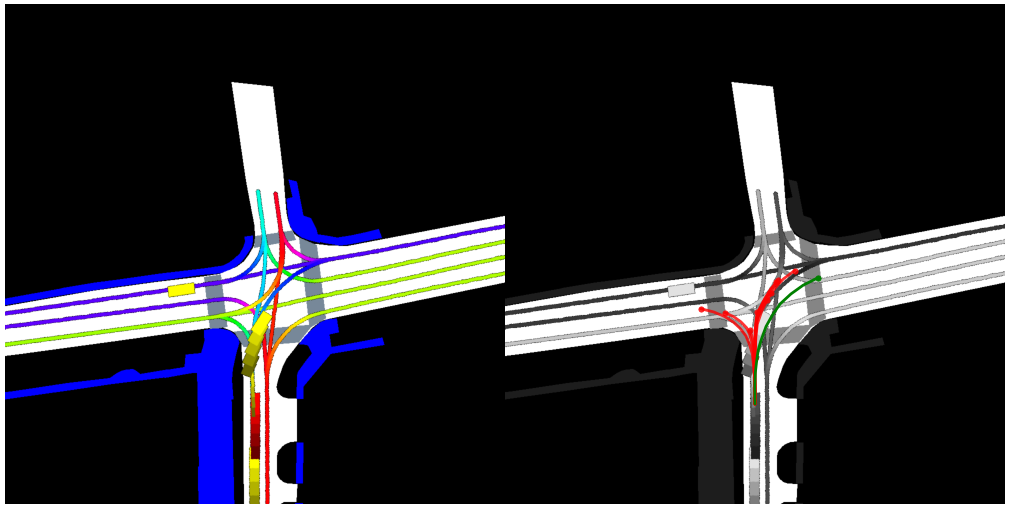}
    \caption{The qualitative comparison of the prediction results on the nuScenes~\cite{caesar2020nuscenes} validation set. From left to right columns: \textbf{a)} traffic situation, \textbf{b)} LAformer without the PnS and HD map information, \textbf{c)} LAformer+PnS that uses the predecessor tracing module to guide the predictions, \textbf{d)} LAformer+HD that uses the HD map information to guide the predictions. The predictions are in red and the corresponding ground truth trajectories are in green.}
    \label{fig:qualitativeresults}
    \vspace{-6mm}
\end{figure*}

\subsection{Qualitative results}
Fig.~\ref{fig:qualitativeresults} illustrates the qualitative results obtained by LAformer, LAformer+PnS, and LAformer+HD.
In the first row, a vehicle is seen traversing an eight-arm intersection. LAformer, without any scene cues, generates a right turn that is incompatible with lane connections, and drives through areas outside lane boundaries. 
In contrast, LAformer+PnS follows traces from predecessors and produces more accurate predictions. With the aid of lane centerlines, LAformer+HD accurately captures the vehicle's driving intent.
The differences among these three models become more apparent in the second row, as the vehicle drives along on the road. 
LAformer generates several predictions that are incompatible with the scene, while the predictions of LAformer+PnS and LAformer+HD are more compliant with the scene.
We also observe that LAformer+PnS generates more divergent predictions in lateral directions in the third row, as it may follow the traces of predecessors turning into different directions in this traffic situation.
Interestingly, in the fourth row, LAformer+PnS predicts a turning modality that is well overlapped with the ground truth trajectory, while LAformer fails to predict the vehicle's intent, and the predictions of LAformer+HD have an offset from the ground truth trajectory. However, we also note that some modalities predicted by LAformer+PnS are not perfectly aligned with the road geometry due to the lack of lane information.

\subsection{Discussion}
\label{sub:discussion}
To avoid false positive predecessor identification, for example, predecessors are too far away from the successor or the angles from the successor to the predecessors are too sharp to make a maneuver, we have tested setting a maximum distance threshold and angle difference. 
This ensures that the identified predecessor agent is spatially close to the successor and driving in a feasible direction. 
However, as a limitation in some cases, the successor may not find any predecessor that meets the criteria, resulting in a lack of observed predecessor information. 
In such cases, the prediction rolls back to conditioning on the past trajectories with an empty predecessor. 
The empirical studies on the nuScenes dataset indicate that even with a few those cases, this did not lead to a significant performance difference (more details in supplementary material).
Overall, the prediction model achieves evidently better performance by including the PnS with the predecessor tracing module.
We leave exploration of more sophisticated solutions, such as knowledge distillation~\cite{zhong2022aware}, to address the cases of no HD map and no predecessors for future work.
\vspace{-3mm}

\section{Conclusion}
\label{sec:conclusion}
In this paper, we propose a probabilistic approach named Predecessor-and-Successor (PnS) to trace the influence of agents (predecessors) identified from the neighboring agents within the same scene on the target agent (successor). 
The moving patterns derived from the traced predecessors serve as informative priors to guide the prediction of the successor's movement.
We use a probabilistic predecessor tracing module to select several highly influential predecessors to account for the stochasticity of the successor's future behavior.
Compared to memory-based models, our method requires no extra effort to collect all the trajectory data beforehand.
Our simple but effective PnS is integrated into a graph-based predictor for pedestrian trajectory prediction on the ETH/UCY datasets, which achieves a new state-of-the-art performance.
Furthermore, we demonstrate the effectiveness of PnS by replacing the HD-map extraction module with PnS in a transformer-based predictor for vehicle trajectory prediction on the nuScenes dataset.
Our method largely mitigates the performance degradation when the map information is removed. 
It also achieves comparable performance when it is compared with a Local Aware module that exploits the historical trajectories traversed in the same location. 

\noindent\textbf{Acknowledgments:} This work is partially funded by MSCA European Postdoctoral Fellowships under the 101062870 - VeVuSafety project.

\appendix
In this supplementary material, we provide further information about the computational performance of the proposed Predecessor and Successor (PnS) method in Section~\ref{sec:performance} and experimental results in Section~\ref{sec:limits}.

\section{Computational performance}
\label{sec:performance}
Table~\ref{tab:computation} demonstrates the computational performance of PnS with the Predecessor Tracing module in terms of model size and inference speed.
It can be seen that LAformer+PnS demonstrates a good inference speed, \eg \SI{22}{ms} for 12 agents and \SI{95}{ms} for 32 agents, which is faster than the data sampling rate of nuScenes (\SI{10}{Hz}).
Also, in the same batch-size setting, LAformer+PnS is much more lightweight in terms of model size and also has a faster inference speed than LAformer+HD. 
This is not surprising because compared with the Predecessor Tracing module, extracting HD map information requires more powerful hidden layers and takes more time to align lane segments with motion dynamics to guide the prediction.
\begin{table}[ht!]
\setlength{\tabcolsep}{5pt}
\centering
\begin{tabular}{l|c|cc}
\toprule
\multirow{2}{*}{Model}    & \multirow{2}{*}{\#Params} & \multicolumn{2}{c}{Inference speed} \\
         &          & Batch size        & Time (ms)       \\ \midrule
LAformer~\cite{liu2023laformer}+PnS & 377K   & 32                & 95             \\
LAformer~\cite{liu2023laformer}+PnS & 377K   & 12                & 22             \\
LAformer~\cite{liu2023laformer}+HD  & 482K   & 12                & 82             \\
LAformer~\cite{liu2023laformer}+HD  & 482K   & 32                & 215             \\

\bottomrule
\end{tabular}
\caption{Model size and inference speed.}
\label{tab:computation}
\end{table}

\section{Further analysis of experimental results}
\label{sec:limits}
\noindent
\textbf{Predecessor Tracing.} To further analyze the efficacy of the Predecessor Tracing module, we evaluate the performance difference by adding distance and angle thresholds in the process of identifying potential predecessors, as shown in Table~\ref{tab:resultsonnuscenes1}. 
\begin{table}[bpht!]
\centering
\begin{tabular}{l|cc}
\toprule
Threshold & mADE$_{10}$ &  mFDE$_{10}$ \\ \midrule
- & 1.21  &  2.32  \\
$\surd$ & 1.21 & 2.34\\ \bottomrule
\end{tabular}
\caption{Thresholding the distance and heading angles for identifying potential predecessors. According to this configuration, the candidate predecessors are identified within \SI{20}{m} $L_2$ distance and $[-90, 90]$ degrees in the field-of-view of the successor. Other agents outside this area are ignored.} 
\label{tab:resultsonnuscenes1}
\end{table}

\begin{figure}[t!]
    \centering
    \includegraphics[clip=true, trim=0pt 0pt 0pt 0pt, width=0.475\textwidth]{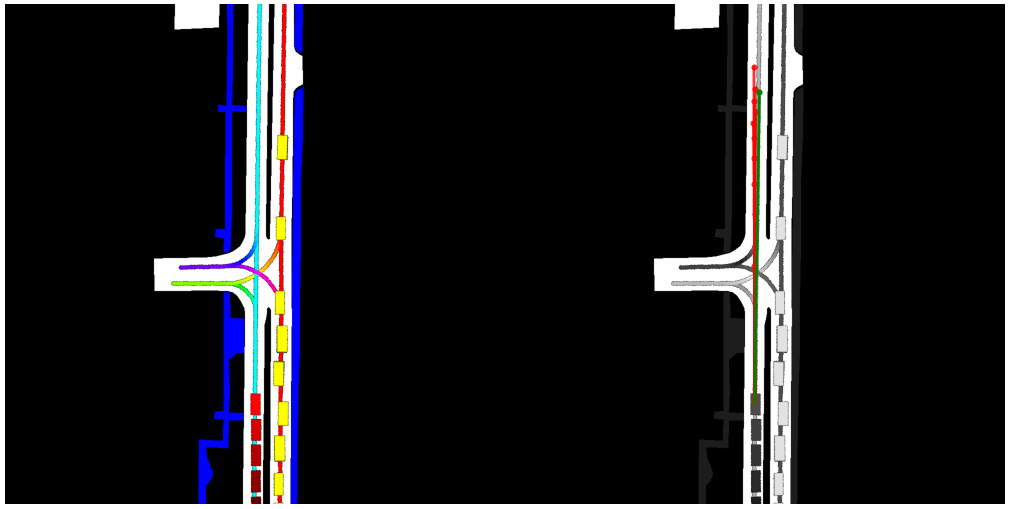}
    \includegraphics[clip=true, trim=0pt 0pt 0pt 0pt, width=0.475\textwidth]{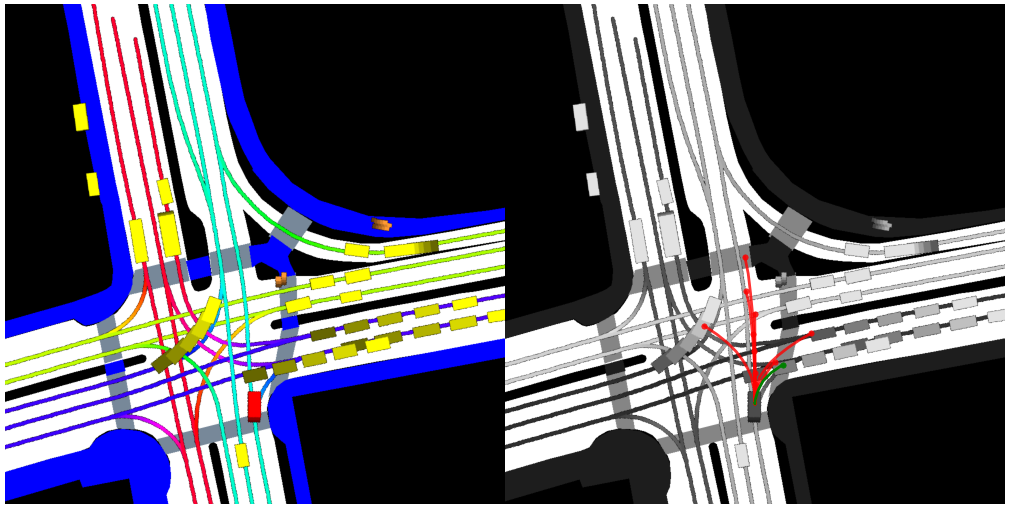}
    \includegraphics[clip=true, trim=0pt 0pt 0pt 0pt, width=0.475\textwidth]{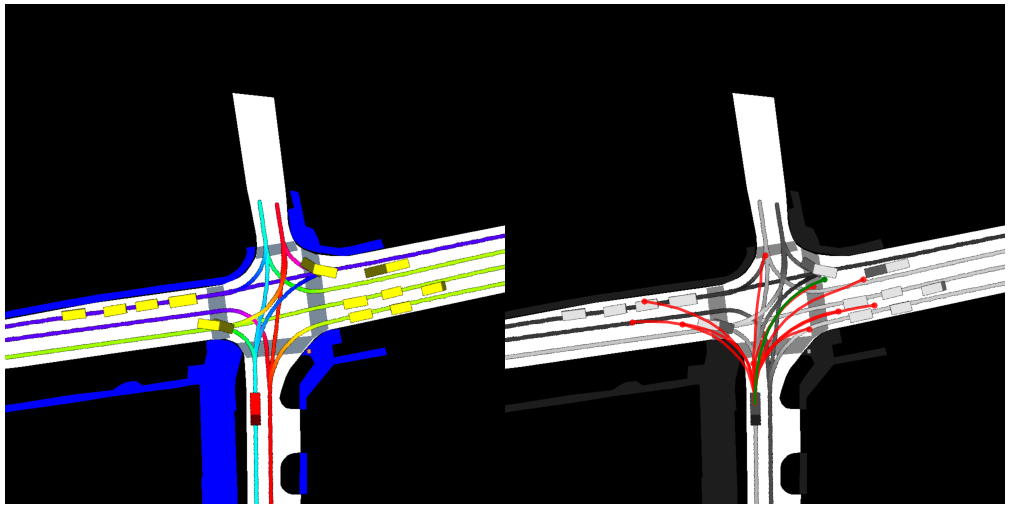}    
    \includegraphics[clip=true, trim=0pt 0pt 0pt 0pt, width=0.475\textwidth]{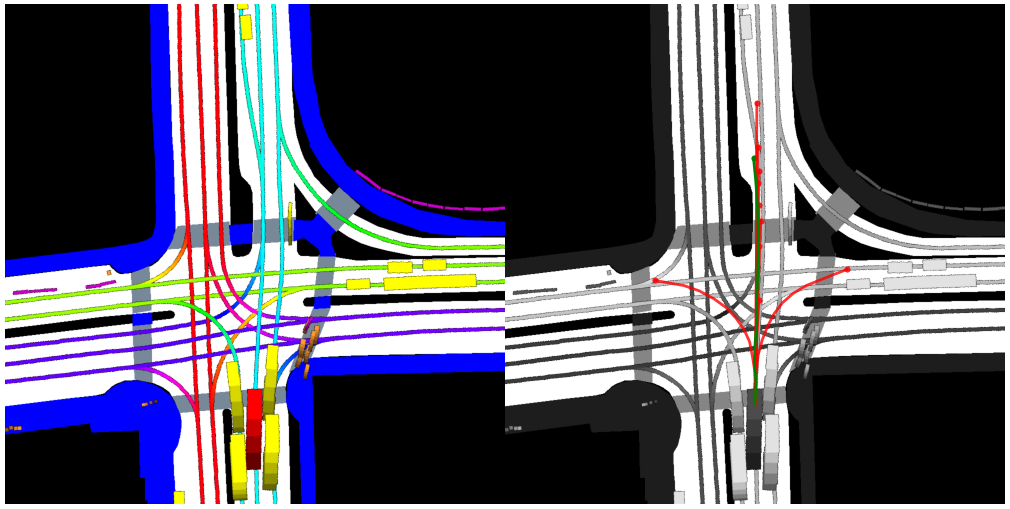} 
    \caption{Challenging cases where LAformer+PnS has difficulties in following the lane connections. Following \cite{deo2022multimodal}, we adopt the same visualization scheme to present the traffic situations on the left column and the corresponding predictions on the right column. The predictions are in red and the corresponding ground truth trajectories are in green.}
    \label{fig:qualitativeresultsappend}
\end{figure}

Specifically, we only search for potential predecessors within \SI{20}{m} in $L_2$ distance and $[-90, 90]$ degrees in the field-of-view of the successor, while ignoring other agents outside this area. 
This setting ensures that the identified predecessor agents are spatially close to the successor and driving in a feasible direction, thus avoiding false positive predecessor identification. 
However, only a marginal performance difference measured in mFDE$_{10}$ is found by adding these thresholds. 
As we mentioned in the main paper, when there is no agent within this area satisfying these thresholds, the prediction rolls back to conditioning on the past trajectories with an empty predecessor.
Nevertheless, distant agents or agents with large angles may also carry benefits information.
For instance, forward driving agents can indicate the driving direction of the lane even these agents are far away from the successor, and an agent may change to the opposite lane after making a U-turn. 
These thresholds may result in a reduced recall of the true predecessors.
Based on this observation we assume that the lack of observations and reduced recall due to the thresholds may contribute to the performance difference. 

\vspace{2mm}
\noindent
\textbf{Challenging cases.}
Figure \ref{fig:qualitativeresultsappend} presents examples of challenging cases where LAformer+PnS faces difficulties with scene constraints. 
Due to the limited observation of predecessors, LAformer+PnS cannot always accurately align its multimodal predictions in accordance with lane connections and driving directions. 
For instance, in the first-row scenario, the model predicts only straight-forward driving, despite through traffic and left-turn traffic sharing a significant segment of the same lane. 
This is because few vehicles were observed for the left turn. 
In the other three scenarios, even though at least one predicted trajectory overlaps well with the corresponding ground truth trajectory, some of the predicted trajectories are not feasible in terms of driving directions and lane connections. 
This indicates that in some cases, it is not sufficient to estimate scene constraints based only on the predecessors' trajectories.

Overall, as illustrated in Table 3 in the main paper, the inclusion of PnS with the Predecessor Tracing module significantly improves prediction accuracy when compared to the map-less setting. 
We envision that our Predecessor Tracing module can inspire the development of more sophisticated techniques for identifying predecessors to address the increased challenges in trajectory prediction tasks that lack explicit scene information. 
We defer the exploration of more sophisticated solutions, such as knowledge distillation~\cite{zhong2022aware}, for addressing scenarios involving neither HD maps nor predecessors, to future work.

{\small
\bibliographystyle{ieee_fullname}
\bibliography{mybib}
}

\end{document}